\documentclass{article}

\PassOptionsToPackage{numbers}{natbib}

\usepackage[final]{neurips_2023}

\usepackage[utf8]{inputenc} % allow utf-8 input
\usepackage[T1]{fontenc}    % use 8-bit T1 fonts
\usepackage{hyperref}       % hyperlinks
\usepackage{url}            % simple URL typesetting
\usepackage{booktabs}       % professional-quality tables
\usepackage{amsfonts}       % blackboard math symbols
\usepackage{nicefrac}       % compact symbols for 1/2, etc.
\usepackage{microtype}      % microtypography
\usepackage{xcolor}         % colors

\usepackage{array, multirow}
\usepackage{adjustbox}
\usepackage{graphicx}
\usepackage{amsmath}
\usepackage{amssymb}
\usepackage{mathtools}
\usepackage{amsthm}
\usepackage{enumitem}
\usepackage{algorithm}
\usepackage{algorithmic}
\usepackage{float}
\usepackage{wrapfig}

\title{Efficient Online Data Mixing For Language Model Pre-Training}

\author{
  Alon Albalak$^1$
  \quad
  Liangming Pan$^1$
  \quad
  Colin Raffel$^{2,3}$
  \quad
  William Yang Wang$^1$ \\
  $^1$University of California, Santa Barbara\\ $^2$University of Toronto\\ $^3$Vector Institute
}

\begin{document}

\maketitle

\begin{abstract}
    The data used to pretrain large language models has a decisive impact on a model's downstream performance, which has led to a large body of work on data selection methods that aim to  automatically determine the most suitable data to use for pretraining.
    Existing data selection methods suffer from slow and computationally expensive processes, a problem amplified by the increasing size of models and of pretraining datasets.
    % Methods in data selection take the approach of finding the individual data points that satisfy certain criteria, but can be slow and computationally expensive, a problem amplified by the ever increasing quantities of pre-training data and model sizes.
    % Another approach is data mixing, which groups datapoints together and determines sampling probabilities across groups, reducing computational complexity. However, data mixing proportions are generally determined prior to training and remain static without taking training dynamics into account.
    Data mixing, on the other hand, reduces the complexity of data selection by grouping data points together and determining sampling probabilities across entire groups. However, data mixing proportions are typically fixed before training and therefore cannot adapt to changing training dynamics.
    To address these limitations, we develop an efficient algorithm for Online Data Mixing (\textbf{ODM}) that combines elements from both data selection and data mixing.
    Based on multi-armed bandit algorithms, our online approach optimizes the data mixing proportions during training.
    Remarkably, our method trains a model that reaches the final perplexity of the next best method with 19\% fewer training iterations, and improves performance on the 5-shot MMLU benchmark by 1.9\% relative accuracy, while adding negligible wall-clock time during pretraining.
\end{abstract}

\begin{figure*}[h]
    \centering
    \includegraphics[width=0.95\textwidth]{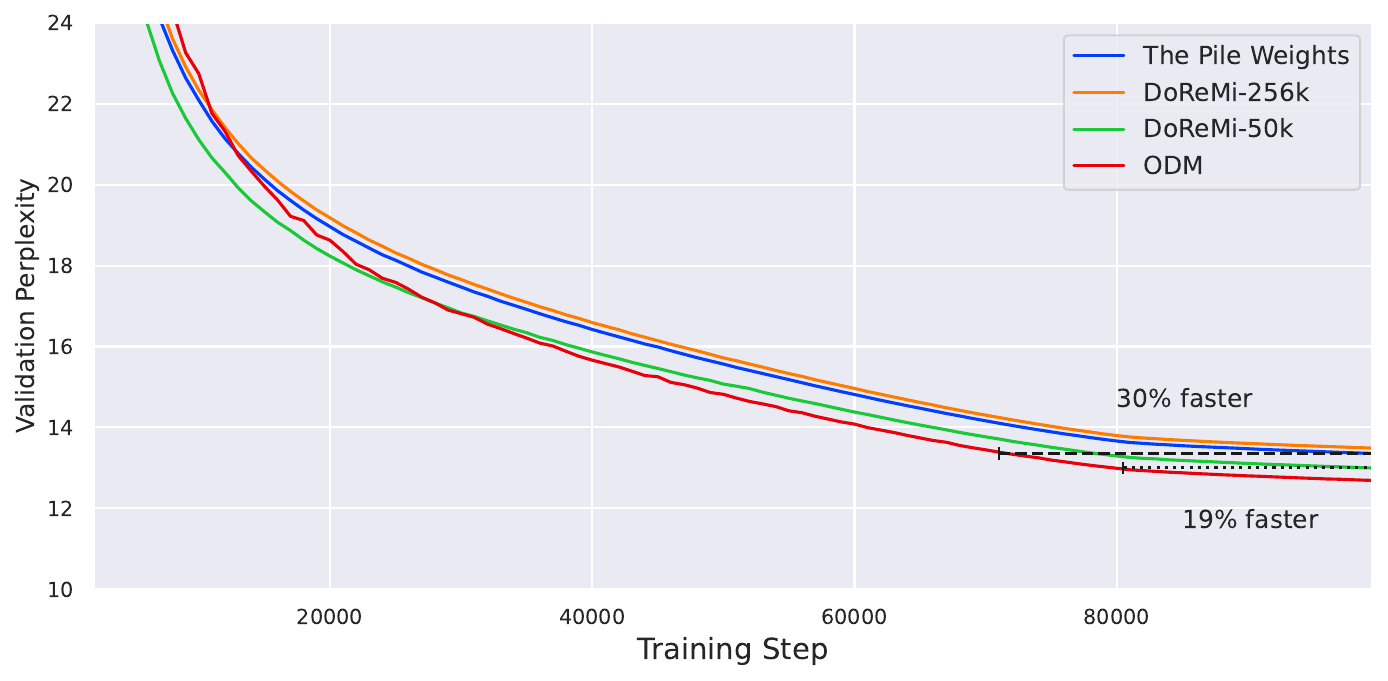}
    \caption{\textbf{Validation perplexity}, unweighted average over 22 domains from The Pile~\cite{gao2020pile}.}
    \label{fig:val_ppl}
\end{figure*}

\section{Introduction}

\label{sec:intro}
% What are we trying to do and why is it relevant?
It is well-known that the training data for machine learning models has a significant influence on their performance.
In particular, the data used to pretrain large language models (LLMs) can be a major factor in the performance of a given LLM.
For example, the 28 different 7-billion parameter models on the Open LLM Leaderboard\footnote{\href{https://huggingface.co/spaces/HuggingFaceH4/open_llm_leaderboard}{Open LLM Leaderboard} accessed on 10/02/2023, 28 models includes only pretrained models without fine-tuning, instruction-tuning, or RL-tuning.} have scores varying from 34.92 to 56.26 even though they all use nearly the same model architecture and training process~\cite{open-llm-leaderboard}. 
It is a widely accepted view that \textit{pretraining is performed so that models can absorb large quantities of information}~\cite{NEURIPS2020_1457c0d6,kirstain-etal-2022-examples,zhou2023lima}, and later training stages such as target task fine-tuning~\cite{albalak2023improving}, instruction fine-tuning~\cite{wei2022finetuned}, and RLHF~\cite{ziegler2020finetuning} primarily refine the model for a specific purpose. This perspective raises the important question of how best to choose pretraining data for training LLMs.
% However, determining what data has high quantities of information is challenging.

% What do previous approaches look like?
Language models are generally trained on data collected from a variety of domains in hopes that data diversity will lead to a higher-quality model, but the data mixing strategy to use (i.e.\ how frequently to sample data from each domain) during training is an open question. For example, when introducing The Pile~\cite{gao2020pile} dataset (consisting of data from 22 domains), the authors suggest higher sampling weights on academic texts and those domains that they felt would provide high-quality data, but these weights are determined using intuition and heuristics, raising the question as to whether a more performant set of weights could be found. The recently proposed DoReMi algorithm~\cite{xie2023doremi} was specifically designed to automatically determine a data mixing strategy for LLM training. DoReMi optimizes domain weights that maximize the information gained of a ``proxy'' model over a ``reference'' model, but requires training multiple models, reducing the method's efficiency. Additionally, we show in this work that their sampling weights don't transfer well across models and thus requires training new ``reference'' and ``proxy'' models in order to determine the best weights for each new model architecture or tokenizer.
These additional steps and considerations reduce the effective efficiency of DoReMi and further increase the already expensive cost of training large language models.
% Feels a bit like the sentence below is almost a different point
Furthermore, both DoReMi and The Pile fix weights throughout training and therefore cannot adapt to changing dynamics over the course of pretraining.

% How do we solve the problem?
In this work, we follow the principle that the best data to train on is the data that maximizes information gained and that a data selection method should introduce negligible computational overhead. Motivated by recent uses of multi-armed bandits (MAB) for auxiliary data selection in few-shot LLM fine-tuning~\cite{albalak2023improving}, we view each data domain as the arm of an MAB and design an algorithm that optimizes the data mixing distribution in an online fashion, thereby adapting to changing training dynamics. Recalling from information theory that perplexity can be thought of as a measure of model uncertainty and the expected information gain from learning the next token, we aim to increase the mixing ratio for domains with the most information to be learned. We therefore utilize the training loss per domain as a reward for our multi-armed bandit algorithm, which fortuitously requires minimal overhead to compute.

To empirically validate the effectiveness and efficiency  our approach, we perform language model pretraining using a 1-billion parameter model trained on 50 billion tokens from the 22 domains found in The Pile~\cite{gao2020pile}. We compare our method with three baseline data mixing methods, finding that our online data mixing algorithm is the most effective, reaching the final validation perplexity of the next best method with 19\% fewer iterations (Figure~\ref{fig:val_ppl}) and improving on 5-shot MMLU~\cite{hendryckstest2021} performance by 3\% relative accuracy over the baseline (Table~\ref{tab:5shot_mmlu}).
Additionally, we find that our method is computationally efficient, introducing a minuscule 0.000007\% overhead.
% with a computation time that varies between 0.0007 and 0.0008 seconds per iteration, a minuscule 70-80 additional seconds over the course of 100,000 training iterations (50 billion tokens).

% main contributions?
% In summary, our contributions are: \textbf{(1)} We propose an efficient online data mixing method and we \textbf{(2)} empirically validate that our method improves in training efficiency and downstream performance over baselines.

\begin{figure*}
    \centering
    \includegraphics[width=0.85\textwidth]{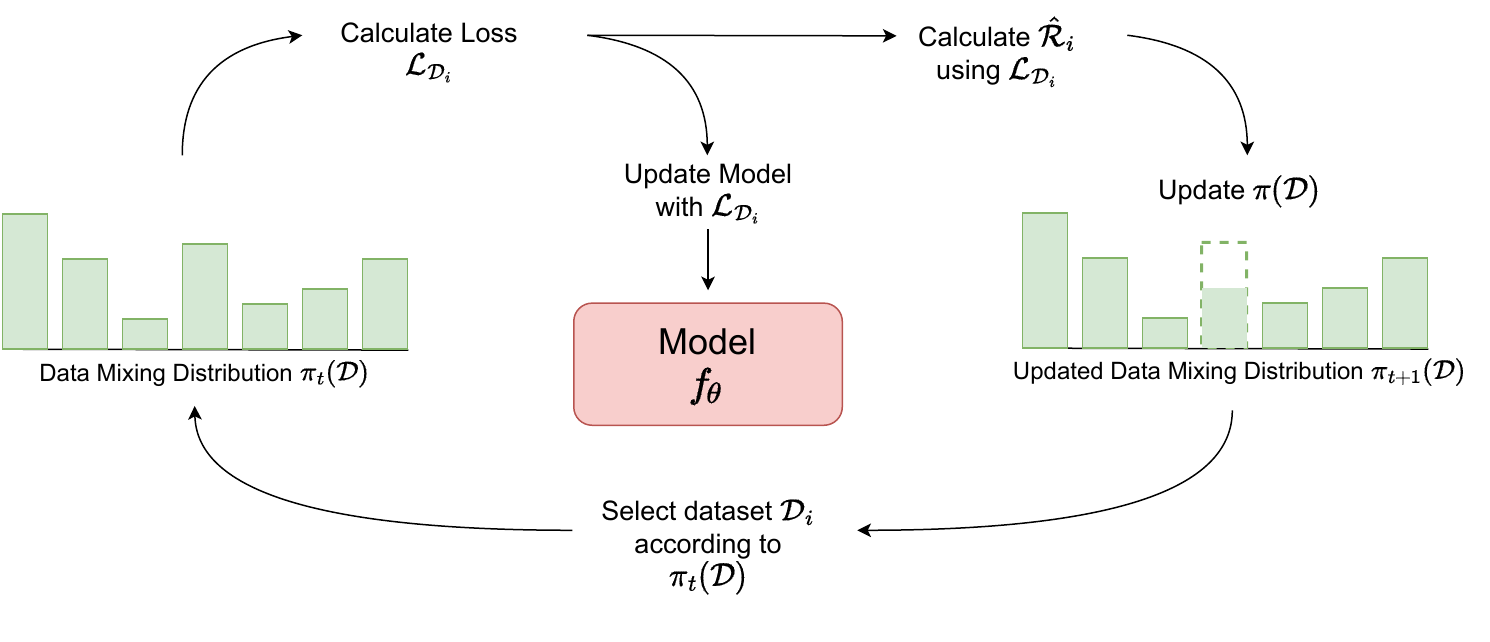}
    \caption{\textbf{Overview of Online Data Mixing (ODM) as a multi-armed bandit}. At each iteration of training, $t$, a dataset $\mathcal{D}_{i}$ is sampled according to the data mixing distribution $\pi$. The loss $\mathcal{L}_{\mathcal{D}_{i}}$ is calculated w.r.t the model $\mathit{f}_{\theta}$ and subsequently used to update the model. Simultaneously, a reward $\hat{\mathcal{R}}_{i}$ is calculated and used to update $\pi$ for the next iteration, $i+1$.}
    \label{fig:odm}
\end{figure*}

\section{Online Data Mixing (ODM)}
In this section, we first define the setting under which online data mixing for language model pretraining takes place (outlined in Figure~\ref{fig:odm}). Then, we formulate the online data mixing problem under the multi-armed bandit (\textbf{MAB}) setting, and describe our reward function which measures information gain and is very efficient to compute. Finally, we describe our algorithm for ODM and present pseudo-code in Algorithm~\ref{alg:main}.

% background info needed to understand our algorithm
\paragraph{Problem setup.}
Consider the setting where we are given $K$ groups of data for language model pretraining, where each group $\mathcal{D}_{i}$ will be sampled according to the probability defined by $\pi(\mathcal{D}_{i})$. Each group $\mathcal{D}_{i}$ could be assigned explicitly according to different domains as in The Pile~\cite{gao2020pile}, or they could determined via some automatic method (as e.g.\ in~\cite{gururangan2023scaling}). In traditional data mixing, each $\pi(\mathcal{D}_{i})$ is fixed prior to training, but in online data mixing, we let each $\pi({\mathcal{D}_{i}})$ be redefined at every training iteration. Given that we want to update $\pi(\mathcal{D}_{i})$ at every training iteration, the problem this work attempts to solve is how to update $\pi({\mathcal{D}_{i}})$ so that the information content of the data being trained on is maximized, and how to do so efficiently.

% Discuss MAB basics: played over turns, each domain is an ``arm'', maintain an exploration rate

\paragraph{Adapting multi-armed bandits to data mixing.}
% Algorithm background
We adopt the multi-armed bandit (\textbf{MAB}) framework to attack the online data mixing problem by formulating it as a Markov decision process~\cite{10.2307/24900506} that is played over $N$ turns.
We design our approach based on Exp3 (\textit{Exp}onential-weight algorithm for \textit{Exp}loration and \textit{Exp}loitation)~\citep{auer2002nonstochastic}. Exp3 defines the policy as a Gibbs distribution based on the empirically determined importance-weighted reward of dataset proportions~\cite{pmlr-v24-seldin12a} and allows for exploration by mixing the Gibbs distribution with a uniform distribution~\cite{auer2002nonstochastic}. Let $\mathcal{E}_{t}$ represent the exploration rate at time step $t$, then the probability of selecting dataset $\mathcal{D}_{i}\in\mathcal{D}$ is defined by $\pi$ as the linear combination of Gibbs and uniform distributions

$
    \pi_{t}(\mathcal{D}_{i}) = (1-K\mathcal{E}_{t})\frac{\exp(\mathcal{E}_{t-1}\hat{R}_{i})}{\sum_{j} \exp(\mathcal{E}_{t-1}\hat{R}_{j})}+\mathcal{E}_{t}
$
where $\hat{R}_{i,t}$ is the moving average of the importance weighted reward
$
    \hat{R}_{i,t} = \alpha\hat{R}_{i,t-1} + (1-\alpha)\frac{R_{i,t}}{\pi_{t-1}(\mathcal{D}_{i})}.
$
We adopt the decaying exploration rate from Seldin et al.~\cite{pmlr-v24-seldin12a}, defined at turn $t$ as
$
    \mathcal{E}_{t} = \min \Bigl\{\frac{1}{K}, \sqrt{\frac{\ln K}{K\cdot t}} \Bigr\}.
$
The main deviation of our method from Exp3 is the use of a moving average estimated reward instead of a cumulative estimated reward. Under normal MAB settings, rewards at each turn are weighted equally, but in our setting we care most about recent rewards. Thus, we still account for past rewards through the use of a moving average, but rewards from the past are weighted less and less moving further into the past.

\paragraph{Designing the reward function.}
% Re-emphasize the interest in instilling maximum information into the model during pre-training as motivation for the loss-based reward.
When designing our reward function we have 2 main goals: (1) ensure that the policy favors data with the highest information content, and (2) minimize the computation required. To achieve these goals, we define the reward to be the current loss for a given dataset group. Formally, at turn $t$, suppose that dataset $\mathcal{D}_{i}$ is sampled from $\pi(\mathcal{D})$, and a batch is sampled as $\{\mathbf{x,y}\}\sim \mathcal{D}_{i}$. Then, the reward is simply $\mathcal{R}_{i,t} = \mathcal{L}(\mathit{f},\mathbf{x,y})$. By formulating the reward as the training loss on a dataset, we add no additional forward or backward passes through the model beyond standard training procedures, minimizing the computation required. Additionally, as discussed in section~\ref{sec:intro}, perplexity (the exponentiated loss) is a measure of expected information gain from each token in a sequence. Thus, by assigning a high reward to datasets with high perplexity, we favor data with the highest information content.

\begin{algorithm}
\caption{Online Data Mixing (ODM)}
\label{alg:main}
\small
\begin{algorithmic}[1]

\REQUIRE $\mathcal{D} = \{\mathcal{D}_{1},\dots,\mathcal{D}_{K}$\}: Grouped dataset
\REQUIRE $\mathit{f}_{\theta}$: Parameterized model
\REQUIRE $\mathcal{L}$: Loss function
\REQUIRE $G$: Gradient accumulation steps

\STATE \textbf{Initialize}: $K=\vert \mathcal{D} \vert$;\quad$\mathcal{E}_{0} = \frac{1}{K}$;\quad$\forall i \in \{1, \dots, K\}: \hat{R}_{i} = 0$

\FOR{$t=1, 2, \dots, N$}

    \STATE $\mathcal{E}_{t} = \min \Bigl\{\frac{1}{K}, \sqrt{\frac{\ln K}{K \cdot t}} \Bigr\}$ \hfill $\rhd$ Update the exploration rate
    
    \STATE $\pi(\mathcal{D}): \pi(\mathcal{D}_{i}) \gets (1-K\mathcal{E}_{t})\frac{\exp(\mathcal{E}_{t-1}\hat{R}_{i})}{\sum_{j} \exp(\mathcal{E}_{t-1}\hat{R}_{j})}+\mathcal{E}_{t}$ \hfill $\rhd$ Calculate the mixing distribution

    % \STATE $\forall \mathcal{D}_{i}\in\mathcal{D}: \pi(\mathcal{D}_{i}) \gets (1-K\mathcal{E}_{t})\frac{\exp(\mathcal{E}_{t-1}\hat{R}_{i})}{\sum_{j} \exp(\mathcal{E}_{t-1}\hat{R}_{j})} \exp(\mathcal{E}_{t-1}R_{i})+\mathcal{E}_{t}$ %\Comment Define the distribution
    \STATE $\forall i=1,2,\dots,K: \mathcal{L}_{\mathcal{D}_{i}} = 0$ \hfill $\rhd$ Reset group losses
    \FOR{$g=1, 2, \dots, G$}
        \STATE Sample $\mathcal{D}_{i}\sim \pi(\mathcal{D})$ and sample a batch $\{\mathbf{x},\mathbf{y}\}$ from $\mathcal{D}_{i}$
        
        \STATE $\mathcal{L}_{\mathcal{D}_{i}} \gets \mathcal{L}_{\mathcal{D}_{i}} + \mathcal{L}(\mathit{f}_{\theta},\mathbf{x},\mathbf{y})$ \hfill $\rhd$ Record group losses for reward updates

    \ENDFOR

    Update model parameters w.r.t $\sum_{i}\nabla_{\theta}\mathcal{L}_{\mathcal{D}_{i}}$

    \FOR{$i \in \{1, \ldots, K\}$ where $\mathcal{L}_{\mathcal{D}_{i}}\neq 0$}
        
        \STATE $\hat{R}_{i} \gets \alpha\hat{R}_{i} + (1-\alpha)\frac{\mathcal{L}_{\mathcal{D}_{i}}}{\pi(\mathcal{D}_{i})}$ \hfill $\rhd$ Update estimated rewards

    \ENDFOR
    
\ENDFOR
\end{algorithmic}
\end{algorithm}

\paragraph{Online data mixing algorithm.}
Our algorithm is shown in pseudocode in Algorithm~\ref{alg:main} and runs as follows: At each turn, the exploration rate $\mathcal{E}_{t}$ is calculated and the policy $\pi$ defines a sampling strategy over all $K$ datasets $\mathcal{D}_{i}\in\mathcal{D}$. Since we are dealing with LLM pretraining which typically uses a large batch size, we assume that we will have $G$ gradient accumulation steps. For each accumulation step we sample one of the datasets $\mathcal{D}_{i}$, then sample a batch $\{\mathbf{x,y}\}\sim \mathcal{D}_{i}$ and calculate the loss $\mathcal{L}_{\mathcal{D}_{i}}$. After accumulating losses, we calculate the gradient w.r.t.\ $\theta$ and update the model. Finally, for each sampled dataset $\mathcal{D}_{i}$, we calculate a reward $\mathcal{R}_{i}$ that is used to update the policy $\pi$ for the next turn. As a practical method to reduce the very high variance of losses at the beginning of language model training, we include a warmup period during which the model trains, but the policy remains stationary. In practice, we find a warmup period of 1\% of total steps to be sufficient.

\section{Experimental Setup}
% Model, training data, testing data
% we use the same setup as the 1B pythia model
\paragraph{Training.}
For our experiments we use The Pile~\cite{gao2020pile}, an 825Gb open-sourced language modelling dataset comprising 22 smaller datasets from various domains including Wikipedia, Github, and PubMed Central. We train decoder-only style transformers using an adapted version of the GPT-NeoX library~\cite{gpt-neox-library}. For all experiments, we train a 1 billion parameter model using the model configuration of Pythia~\cite{biderman2023pythia}. We train using a batch size of 60 sequences per GPU, and accumulate gradients across 8 GPUs in parallel ($G=8$) to reach a total batch size of 480 samples. We let the sequence length be 1024 and pack sequences together~\cite{roberts2022scaling}. We train for a total of 100,000 steps, reaching 50 billion tokens. For ODM, we initialize the domain weights using those defined by The Pile. The full model configuration and hyperparameters can be found in Appendix~\ref{sec:model_config}.

\paragraph{Evaluation.}
To validate the performance of our approach and the baselines, we compute perplexity on held-out validation and test data from each domain of The Pile. Additionally, we evaluate each model on downstream capabilities by performing multiple choice classification on the 57 tasks from MMLU~\cite{hendryckstest2021}. For each task in MMLU we use 5 in-context examples.

\paragraph{Baselines.}
We compare the performance of our method against that of the original domain weights suggested by The Pile~\cite{gao2020pile}, and refer to it as The Pile Weights (\textbf{TPW}). Additionally, we compare with the domain weights proposed by DoReMi~\cite{xie2023doremi}, but empirically find that the weights do not perform as published. However, after discussion with the authors, we attained weights that were re-calculated on the same tokenizer as ours\footnote{It is hypothesized by the authors of \cite{xie2023doremi} that different tokenizers may lead to different domain weights, but is still an open question why that may be the case.}. The original DoReMi weights are computed with a 256k vocabulary tokenizer while we use a 50k vocabulary tokenizer, so to specify each DoReMi baseline we name them \textbf{DoReMi-256k} and \textbf{DoReMi-50k}.

\begin{figure*}
    % \vspace{-5mm}
    \centering
    \includegraphics[width=\textwidth]{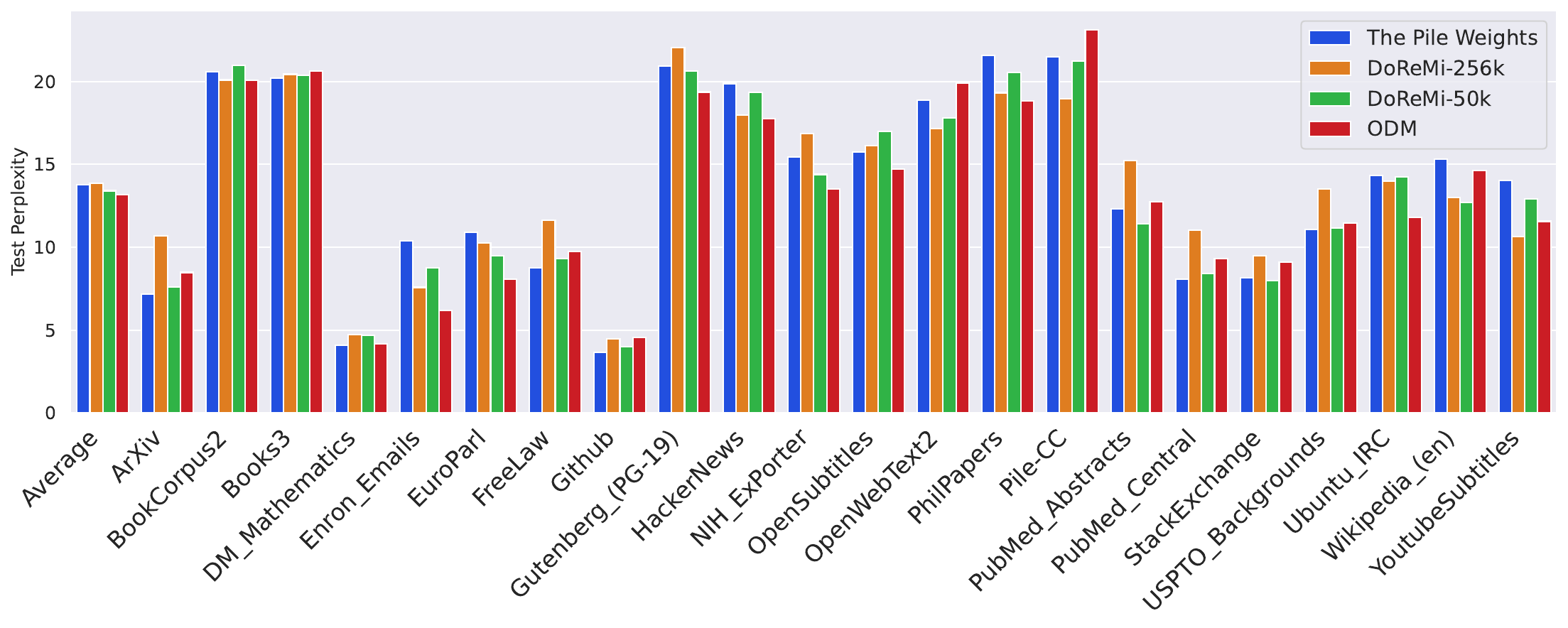}
    \vspace{-5mm}
    \caption{\textbf{Test perplexity} on average, and on 22 individual domains.}
    \label{fig:test_ppl}
    % \vspace{-5mm}
\end{figure*}

% \begin{wrapfigure}{r}{0.5\textwidth}
%     \vspace{-10mm}
%     \centering
%     \includegraphics[width=0.5\textwidth]{figures/mmlu5shot.png}
%     \vspace{-5mm}
%     \caption{Average 5-shot accuracy on MMLU.}
%     \label{fig:5shot_mmlu}
%     % \vspace{-5mm}
% \end{wrapfigure}

\section{Findings and analysis.}
In Figures~\ref{fig:val_ppl} and~\ref{fig:test_ppl} we compare the perplexities of training models using ODM with the baseline data mixing methods. Table~\ref{tab:5shot_mmlu} shows the average 5-shot accuracy on MMLU of ODM and baseline methods.
% , and Figure~\ref{fig:icl_examples_mmlu} shows the effects of varying the number of in-context examples used on MMLU for each of the methods.

\paragraph{Main results.}
% ODM improves on all metrics, on average
Figure~\ref{fig:val_ppl} shows that ODM achieves the final performance of the originally suggested Pile weights (TPW) with 30\% fewer iterations, and 19\% fewer than DoReMi-50k. Additionally, Figure~\ref{fig:val_ppl} shows that ODM's final validation perplexity is 4.8\% lower than TPW, 2.4\% lower than DoReMi-50k, and 4.9\% lower than DoReMi-256k, emphasizing how the DoReMi method is not transferrable across models. These results show that ODM improves the training efficiency compared with static data mixing methods. Additionally, Table~\ref{tab:5shot_mmlu} shows that ODM leads to better downstream performance in 5-shot classification tasks, improving over TPW by 3\%, and DoReMi-50k by 1.9\%.

\begin{wraptable}{r}{0.5\textwidth}
    % \vspace{-3mm}
    \centering
    \small
    \begin{tabular}{c | c }
        \toprule
        \textbf{Method} & \textbf{Accuracy} \\
        \midrule
         The Pile Weights & 0.27469 \\
         DoReMi-256k & 0.27596 \\
         DoReMi-50k & 0.27887 \\
         ODM & 0.28416\\
         \bottomrule
    \end{tabular}
    \caption{\textbf{Average 5-shot accuracy on MMLU}}
    \label{tab:5shot_mmlu}
\end{wraptable}

% Results from average test perplexity
Figure~\ref{fig:test_ppl} shows the test perplexity of each method on held-out data as well as the average perplexity. Surprisingly, we find that the original domain weights reported for DoReMi~\cite{xie2023doremi} (DoReMi-256k) leads to test perplexity that is, on average, 0.7\% worse than The Pile Weights, in direct contradiction with their original findings. However, DoReMi-50k does improve over The Pile Weights by 2.6\%, demonstrating that the domain weights determined by the DoReMi method do not transfer well across models.

% Can we create a takeaway for this paragraph?
% If we show that a very high mixing ratio is required to do well on a task, can we say that there is a tradeoff of doing less well on that task, but that the leftover mixing will go to other domains which in turn do better?
% Not sure exactly how to phrase that tradeoff
\paragraph{The effects of data mixing optimization objectives on individual domain performance.}
% How does each model compare on individual domains
Here we compare the empirical effects of the contrasting optimization of objectives of ODM and DoReMi on individual domains. Recall that the reward function used in ODM favors dataset groups with the greatest information gain (highest loss) at each step, and that DoReMi's objective is to maximize the information gain of a ``proxy'' model over a ``reference'' model (i.e.\ ``minimize the worst-case excess loss''). To see these different objectives in effect, we group the performance of each method into one of three buckets: best, worst, or in the middle, where the ideal method would have all 22 domains in the ``best'' category. Interestingly, we find that The Pile Weights are almost evenly distributed across all 3 buckets, doing worst in 7 domains, best in 7, and in the middle for the remaining 8. As expected from a method that optimizes for the best worst-case scenario, we find that DoReMi-50k's test perplexity is often not the best or the worst, but falls in the middle. In fact, 17/22 domains are in the middle, only performing best on three domains (PubMed\_Abstracts, StackExchange, and Wikipedia\_(en)), and worst on only two domains (BookCorpus2 and OpenSubtitles). On the other hand, using ODM leads to the best perplexity on 9 domains, with 9 more in the middle, and only performing the worst on 4 domains (Books3, Github, OpenWebText2, and Pile-CC). Notably, two of the domains where ODM performs worst are web text domains but this decreased performance does not seem to have a negative impact on downstream performance.

\begin{figure}
    \centering
    \includegraphics[width=0.9\textwidth]{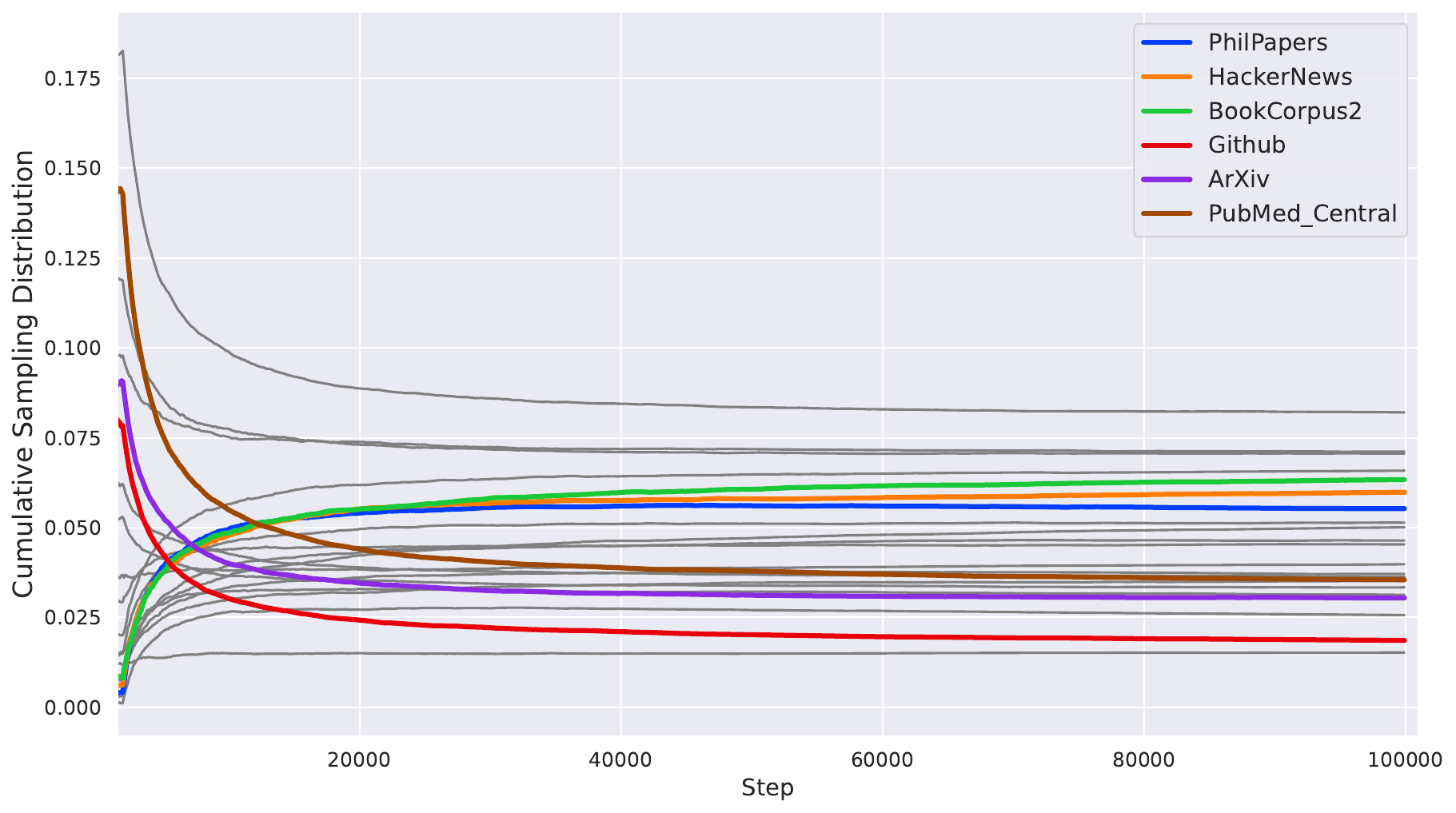}
    \caption{\textbf{The cumulative sampling distribution of ODM} calculated as the samples per domain out of the total number of samples trained on. Highlighted lines are the six domains whose final sampling distributions have increased/decreased the most from initialization.}
    \label{fig:csd}
\end{figure}

\paragraph{What does ODM's sampling policy look like?}
In Figure~\ref{fig:csd} we show the cumulative sampling distribution of each domain over the course of training. Note that ODM is initialized with The Pile Weights, which are the initial values on the left. Figure~\ref{fig:csd} highlights the three datasets whose mixing ratio increased the most (PhilPapers, HackerNews, and BookCorpus2), and the three datasets whose mixing ratio decreased the most (Github, ArXiv, and PubMed\_Central). It is evident from this figure that ODM finds a sampling distribution which is closer to uniform than The Pile Weights. We also see that the distribution for most domains stabilizes early on in training ($\sim40000$ iterations). Beyond the 40000 step, the distribution is still changing, but at a much lower rate. For example, we see that the mixing ratio for Github is still decreasing and the ratio for both BookCorpus2 and HackerNews are increasing all the way until the end of training. 

\paragraph{Why does ODM's validation perplexity start off high?}
% On validation PPL, our model starts off worse. Why is that?
Figure~\ref{fig:val_ppl} shows that although our method outperforms the baselines, at the beginning of training ODM actually has higher perplexity than other methods. We believe that this is due to the homogeneity of the micro-batches used in ODM, whereas other methods see a greater mixture of data in each batch. In preliminary experiments we trialed a version of ODM that uses data from a single domain in all gradient update steps, and found that this exacerbates the phenomena leading to a perplexity that starts even higher. This suggests that one of the weaknesses of our method is the requirement that each batch comes from the same grouped dataset. This problem can be alleviated by decreasing the micro-batch size, but this comes with technical considerations as simply decreasing micro-batch size will reduce GPU utilization, and lead to slower wall clock time. Likely, a better solution would be to mix domains within micro-batches during the warm-up phase, which would lead to validation perplexity exactly the same as The Pile Weights, but gaining the advantages of ODM after the warm-up.

% \paragraph{Limitations.}
% Throughout this work we have motivated the use of training loss as a reward by suggesting that it will instill maximum information into the model. However, this reward will incidentally optimize for any data which has random tokens, and any pretraining method that does not remove such data will suffer.

\section{Conclusion}
The method proposed in this work demonstrates the effectiveness of online data mixing formulated as a multi-armed bandit problem. Additionally, we showed that by designing a reward which attempts to maximize information gain, we can train models that achieve lower perplexity on held-out data in fewer iterations than baseline methods. Furthermore, by utilizing the training loss as a reward, the proposed method is very computationally efficient, adding a trivial 0.000007\% additional wall-clock time to training.

% Future directions: online variant of DoReMi, where the reward takes into account the performance of a reference model

% \section{Notes}
% What do we do differently?
% \begin{itemize}
%     \item We combine ideas from data selection (slow process that selects individual data points either prior to or during training) with data mixing (all computation happens prior to training, has no input from current model) to develop a fast and efficient algorithm for ``online data mixing'' (name can be improved).
%     \item There are two flaws with prior data mixing methods: (1) is that they don't take training dynamics into account (they are a stationary distribution), and (2) is that as new data domains are added to a collection, the methods need to be repeated to account for the new data (existing methods require training many models to empirically determine the best mixing ratio before training the final model, recently DoReMi has reduced this to only training 2 models prior to the final model [but it has other issues as well]). 

% \end{itemize}

% \begin{itemize}
%     \item Overall results - Improved efficiency, better downstream performance
%     \item Specifics - which tasks does our method do better/worse on? Is there a trend there?
%     \item Specifics - which individual domains does our method do better than baselines.
%     \item Plot of domain ppl vs. sampling probability. Theory is that domains that are more frequently sampled will have lower ppl, but is that strictly true? If so, it invalidates DoReMi's findings
% \end{itemize}

\medskip

\bibliography{custom}
\bibliographystyle{unsrt}

%%%%%%%%%%%%%%%%%%%%%%%%%%%%%%%%%%%%%%%%%%%%%%%%%%%%%%%%%%%%

\newpage

\appendix

\section{Model Configuration}
\label{sec:model_config}
Our 1-billion parameter model uses a sequence length of 1024, has 16 layers with a hidden size of 2048, 16 attention heads, and rotary positional embeddings~\cite{su2022roformer}. We use FlashAttention~\cite{dao2022flashattention} to reduce training time. We use the Adam optimizer~\cite{DBLP:journals/corr/KingmaB14} with a linear warmup over 1000 iterations from a minimum learning rate of 2.5e-5 to a maximum learning rate of 2.5e-4, and then decay the learning rate with a cosine schedule down to the minimum of 2.5e-5 again. We use the GPT-NeoX-20B tokenizer~\cite{gpt-neox-20b}.

% \section{Notes for future}
% A similar idea was used in the vision and language area by~\cite{piergiovanni2023dynamic}.

\end{document}